\documentclass[11pt]{article}

\usepackage[final]{acl}

\usepackage{times}
\usepackage{latexsym}
\usepackage[T1]{fontenc}
\usepackage[utf8]{inputenc}
\usepackage{microtype}
\usepackage{inconsolata}
\usepackage{graphicx}
\usepackage{multirow}
\usepackage{booktabs}
\usepackage{amsmath}
\usepackage{amssymb}
\usepackage{xurl}

\hypersetup{
  pdftitle={Beyond Accuracy: Robustness, Cost, and Governance Trade-offs for Vision-Language Models in Templated Document Extraction},
  pdfauthor={Kushal Patel, Pushkal Shrivastava, Mackenzie Lees, Qirui Lu, Bhargobjyoti Saikia, Liying Li, Junlin Jiang}
}

\title{Beyond Accuracy: Robustness, Cost, and Governance Trade-offs for
Vision-Language Models in Templated Document Extraction}

\author{
  Kushal Patel, \; Pushkal Shrivastava, \; Mackenzie Lees, \; Qirui Lu \\
  \textbf{Bhargobjyoti Saikia, \; Liying Li, \; Junlin Jiang} \\
  John Hancock, 200 Berkeley St, Boston, MA 02116, USA \\
  \texttt{\{kpatel, pxshrivastava, mlees, qirui\_lu\}@jhancock.com} \\
  \texttt{\{bsaikia, liyingli, junlin\_jiang\}@jhancock.com}
}

\begin{document}
\maketitle

\begin{abstract}
Vision-language models (VLMs) are increasingly used to extract
structured fields from business documents, yet most evaluations report
accuracy on clean benchmarks and offer little guidance to practitioners
choosing an approach for a given task complexity. We address this gap
with a measurement-grounded study and an open-source release. Across
eleven systems (three commercial, two reasoning, five open-source VLMs
in pretrained and fine-tuned form, and a non-LLM OCR$\to$regex floor)
scored on a 750-document held-out pool of synthetic checks, fine-tuning
on 3K samples lifts the best open-source VLMs above F1 0.98---above
every zero-shot commercial system on this task---while GPT-5 leads the commercial
pool on F1 and Claude Sonnet 4.5 collapses on Date. To turn these measurements
into actionable choices, we introduce a practitioner-oriented
selection framework that maps a task profile (quality, latency,
governance, volume) to a recommended approach via filtering and
total-cost minimization, illustrated on a hypothetical mid-volume
document-extraction scenario.%
\footnote{\textbf{Disclaimer.} This paper reports an experimental
evaluation on synthetic data. It does not represent deployment
guidance, procurement guidance, or the views of any corporation.
No customer documents, internal company metrics, or non-public
operational data were used in any experiment or in the illustrative
scenario of Section~\ref{sec:case}.}
\end{abstract}

\section{Introduction}

Intelligent document processing (IDP) underpins high-volume operations in
finance, healthcare, and insurance. Recent vision-language models (VLMs)
promise end-to-end field extraction without explicit OCR
\citep{bai2025qwen2_5vl,marafioti2025smolvlm}, and large multimodal LLMs
\citep{openai2023gpt4} approach human-level accuracy on clean documents.
For a practitioner facing a new extraction task, however, public
leaderboards offer little actionable guidance: they report accuracy on
curated, well-aligned scans and rarely describe the joint operating
conditions---scan-defect robustness, latency budget, deployment governance,
and processing volume---that determine which approach is feasible at
production.

We focus on structured field extraction from templated business forms,
a prevalent and high-volume class of enterprise document processing
tasks. Our primary study is on bank checks; we additionally report
a cross-industry generalization test on insurance application forms.

Our central question is operational: \emph{given a target task with a
specified complexity (quality threshold, latency budget, governance
requirement, expected volume), which class of approach should a
practitioner adopt?} We answer it through releasable artifacts and a
selection framework grounded in measurement. Because frontier model
lineups change on a quarterly cadence, our central contribution is
the \emph{evaluation-and-selection framework} rather than any fixed
vendor ranking: the model pool reported here is a snapshot, and new
entrants can be evaluated by re-populating the candidate set
$\mathcal{M}$ (Section~\ref{sec:framework}) with updated
measurements from the released harness.
Concretely, we make four contributions.

\paragraph{Contributions.}
(i)~A \textbf{check-extraction dataset} of 3{,}748 synthetic images
with four-field ground truth (Payee, Amount, Date, Bank).
(ii)~An \textbf{open-source code release} covering data generation,
LoRA and full fine-tuning, prompt and API harnesses, and per-sample
diagnostics.%
\footnote{The dataset~(i) and code release~(ii) are available at
\url{https://github.com/manulife-ai/beyond-accuracy-vlm-emnlp2026/};
see Appendix~\ref{app:repro} for reproducibility details.}
(iii)~An \textbf{empirical evaluation} of eleven systems---three
commercial (GPT-4.1 vision, OCR+GPT-4.1, Azure Content Understanding),
two reasoning (GPT-5, Claude Sonnet 4.5), five open-source VLMs
(Qwen2.5-VL-3B/7B, SmolVLM2-2.2B/500M/256M, each in pretrained and
fine-tuned form), and a non-LLM OCR$\to$regex floor---with per-field
and per-error-mode breakdowns over $>$45K field-level predictions.
(iv)~A \textbf{practitioner-oriented selection framework} mapping a
task profile to a recommended approach via governance/quality/latency
filters and total-cost minimization, instantiated on an illustrative
mid-volume document-extraction scenario.

\section{Related Work}

\paragraph{VLMs for document understanding.}
End-to-end models such as Donut \citep{kim2022donut} and document-tuned
multimodal LLMs eliminate explicit OCR. Recent open-source VLMs
\citep{bai2025qwen2_5vl,marafioti2025smolvlm} target on-device inference but have
not been systematically evaluated on enterprise document distortions.

\paragraph{Robustness in document AI.}
Most document benchmarks (FUNSD~\citep{jaume2019funsd},
CORD~\citep{park2019cord}, RVL-CDIP~\citep{harley2015rvlcdip}) provide
clean scans, and prior OCR-robustness work has tended to address
individual preprocessing axes in isolation
\citep[e.g., binarization under non-uniform illumination,][]{michalak2020robust}
in the pre-VLM era~\citep{cui2021documentai}. The robustness of modern VLMs
to \emph{compound} enterprise distortions---perspective, rotation, and
noise---on structured-extraction tasks remains underexplored.

\paragraph{Deployment trade-offs.}
Cost--accuracy Pareto analyses for LLM APIs \citep{chen2023frugalgpt} have
addressed text-only inference. Recent work in enterprise document
processing has begun to expose the same trade-off space: hybrid OCR+LLM
routing across 25 configurations for copy-heavy identity documents
\citep{wang2025hybrid}, and multi-agent orchestration benchmarks on
10\,K SEC filings that report cost-per-document and Pareto positions
alongside accuracy \citep{kulkarni2026multiagent}. Our work extends
this analysis to multimodal field extraction and treats governance as
a configurable task parameter: a given workload may or may not impose
constraints (legal, policy, contractual, or operational) that narrow
the set of admissible deployment options, and the framework simply
encodes whichever constraints the practitioner specifies.

\section{Experimental Setup}
\label{sec:setup}

\paragraph{Task.}
Models are prompted to return a JSON object with four fields per check:
\textbf{Payee} (free-text), \textbf{Amount} (numeric), \textbf{Date} (MM/DD/YYYY),
and \textbf{Bank} (categorical: Chase, Citi, BOA). All variants share an
identical prompt and schema.

\paragraph{Dataset.}
We render 3{,}748 synthetic check images from a structured table of field
values using a template-driven pipeline with controlled appearance
perturbations. Of these, 2{,}998 are used for fine-tuning and 750
form the held-out test pool (identical for every model in this
study); no document appears in both splits.

\paragraph{Train/test disjointness.} Because the dataset is
template-driven, leakage is a legitimate concern. We split at
the row level of the underlying field table before rendering: the Payee,
Amount, and Date values in the 750-document test pool are drawn from rows
disjoint from the 2{,}998-document training table (random partition,
seed 42, fixed for all experiments). Bank, being a 3-way categorical, is
shared by construction; its near-perfect accuracy across all systems
(Section~\ref{sec:perfield}) should therefore be read as classification
on a closed vocabulary rather than evidence of generalisation. Template
backgrounds are also held disjoint: the rendering pipeline draws from
separate background pools for train and test.

\paragraph{Models.}
We evaluate three commercial systems---GPT-4.1 (vision-only),
OCR+GPT-4.1 (Azure Document Intelligence + GPT-4.1), and Azure Content
Understanding (ACU)---two reasoning models---GPT-5 and Claude Sonnet
4.5---five open-source VLMs: Qwen2.5-VL-3B and 7B (LoRA fine-tuned,
$r{=}16$, $\alpha{=}32$, 1.08\% / 1.23\% trainable parameters), and
SmolVLM2-256M, SmolVLM-500M, SmolVLM2-2.2B (full fine-tuning)---and a
non-LLM OCR$\to$regex baseline (Azure Document Intelligence Read +
hand-written field-specific patterns) that bounds the floor of what a
zero-LLM-cost pipeline achieves on this task.
Open-source models are evaluated both pretrained and after fine-tuning.
Fine-tunes use 2{,}998 examples, effective batch size four, paged
AdamW-8bit; Qwen-3B and the three SmolVLM variants run 3 epochs, with
learning rates $2{\times}10^{-4}$ (Qwen-3B LoRA) and $1{\times}10^{-5}$
(SmolVLM full fine-tuning). Qwen-7B uses 5 epochs at $1{\times}10^{-4}$
LoRA learning rate, a longer/gentler schedule that keeps the 7B adapter
from truncating long outputs.

\paragraph{Scope of baselines.} We do not evaluate purpose-built
structured-extraction encoders (Donut~\citep{kim2022donut},
LayoutLMv3~\citep{huang2022layoutlmv3}, Pix2Struct~\citep{lee2023pix2struct},
UDOP~\citep{tang2023udop}) in this release: our framework requires
candidates that produce structured JSON directly, whereas these
models require task-specific decoding heads. Recent 2025 work on
document key information extraction~\citep{wang2025marten,yu2025docthinker}
benchmarks task-adapted VLMs against other VLMs rather than
LayoutLMv3/Donut-era models, reflecting a community shift in what
constitutes a competitive extraction system; our candidate set is
aligned with this current practice. Incorporating purpose-built
encoder-decoder models is future work.

\paragraph{Metrics.}
We report entity-level micro-F1 with TP/FP/FN aggregated over all four fields,
per-field F1, document-level accuracy (all four fields correct), JSON parse
success rate, and end-to-end latency. We additionally analyze per-sample debug
logs to categorize errors as \emph{under-extraction} (empty prediction),
\emph{wrong value}, or \emph{parse failure}.

\section{Results}
\label{sec:results}

\subsection{Overall Performance}

Table~\ref{tab:main} reports headline metrics. All models---commercial,
reasoning, and open-source---are scored on the same 750-document
held-out PNG test pool.

\begin{table}[t]
\centering\small
\setlength{\tabcolsep}{4pt}
\begin{tabular}{lcccc}
\toprule
Model & F1 & DocAcc & Lat. (s) & JSON Fail \\
\midrule
\multicolumn{5}{l}{\textit{Commercial \& reasoning}} \\
GPT-5                & \textit{0.928} & 41.7  & 26.3 & 0.0\% \\
OCR+GPT-4.1          & 0.917 & 40.7  & 18.1 & 0.0\% \\
GPT-4.1 (vision)     & 0.909 & 45.4 & 12.2 & 0.0\% \\
ACU                  & 0.889 & 33.1  & 12.8 & 0.0\% \\
Claude Sonnet 4.5    & 0.857 & 14.8  & 8.5  & 0.0\% \\
\midrule
\multicolumn{5}{l}{\textit{Non-LLM floor}} \\
OCR+Regex            & 0.395 & 0.0   & 8.6  & --- \\
\midrule
\multicolumn{5}{l}{\textit{Open-source, fine-tuned}} \\
Qwen2.5-VL-7B  & \textbf{0.985} & \textbf{88.9} & 3.53 & 0.0\% \\
Qwen2.5-VL-3B  & 0.983 & 88.4 & 4.18 & 0.0\% \\
SmolVLM2-2.2B  & 0.906 & 46.7 & 1.47 & 0.0\% \\
SmolVLM-500M   & 0.854 & 45.1 & \textbf{0.96} & 5.3\% \\
SmolVLM2-256M  & 0.726 & 22.4 & 1.62 & 30.7\% \\
\midrule
\multicolumn{5}{l}{\textit{Open-source, pretrained (no FT)}} \\
Qwen2.5-VL-7B  & 0.921 & 50.7 & 2.53 & 0.0\% \\
Qwen2.5-VL-3B  & 0.847 & 24.5 & 4.75 & 0.0\% \\
SmolVLM2-2.2B  & 0.504 & 0.4  & 2.03 & 3.3\% \\
SmolVLM-500M   & 0.000 & 0.0  & 1.80 & 100\% \\
SmolVLM2-256M  & 0.000 & 0.0  & 1.30 & 100\% \\
\bottomrule
\end{tabular}
\caption{Headline metrics on the 750-document held-out PNG test pool
(identical pool for every model). Fine-tuned rows use a fixed
adaptation budget of 2{,}998 training documents (LoRA for Qwen,
full fine-tuning for SmolVLM; recipes in Section~\ref{sec:setup}).
\textbf{Bold} marks the best in column among open-source FT;
\textit{italic} marks the best F1 within the commercial / reasoning
block. Open-source latencies are
measured on a single NVIDIA A100 80 GB PCIe (bf16, batch size 1,
five-document warmup discarded); see Section~\ref{sec:gpu} for the
A100$\to$H100 sensitivity analysis. The OCR+Regex row is a
non-LLM floor baseline (Azure Document Intelligence Read + hand-written
field-specific patterns); its outputs are field-keyed dictionaries, not
JSON, so the JSON-failure column is not applicable.}
\label{tab:main}
\end{table}

Within the commercial pool, GPT-5 leads on F1 narrowly above
OCR+GPT-4.1 and GPT-4.1 vision but at $2{\times}$ the latency of the
latter, while GPT-4.1 vision has the highest document-level accuracy;
ACU is the most balanced across fields (Section~\ref{sec:perfield}),
and Claude Sonnet 4.5 trails, dragged down almost entirely by Date.

Under a fixed adaptation budget of 2{,}998 training documents
(Section~\ref{sec:setup}), the best fine-tuned open-source models
exceed every zero-shot commercial system on this schema---Qwen2.5-VL-7B
FT reaches 0.985 and Qwen2.5-VL-3B FT 0.983
(within 0.002 of each other; Appendix~\ref{app:variance} confirms
this gap is within seed noise), both with zero JSON parse failures,
suggesting the 3B adapter already saturates the available signal on
this schema. Even SmolVLM2-256M becomes usable with fine-tuning,
moving from 0.000 to 0.726 F1 and from 100\% to 30.7\% JSON parse
failure---operationally fragile but viable for clean-document pilots.
The non-LLM OCR$\to$regex floor trails every commercial and
fine-tuned system (Tables~\ref{tab:main}--\ref{tab:perfield}).

\subsection{Per-Field Analysis}
\label{sec:perfield}

\begin{table}[t]
\centering\small
\setlength{\tabcolsep}{4pt}
\begin{tabular}{lcccc}
\toprule
Model & Payee & Amount & Date & Bank \\
\midrule
\multicolumn{5}{l}{\textit{Commercial \& reasoning}} \\
OCR+GPT-4.1       & \textbf{0.874} & \textbf{0.994} & 0.769 & \textbf{1.000} \\
GPT-4.1 (vision)  & 0.856 & 0.963 & 0.791 & \textbf{1.000} \\
ACU               & 0.780 & 0.960 & \textbf{0.795} & 0.988 \\
GPT-5             & 0.870 & 0.970 & 0.772 & 0.999 \\
Claude S.~4.5     & 0.810 & 0.874 & \textit{0.468} & 0.993 \\
\midrule
\multicolumn{5}{l}{\textit{Non-LLM floor}} \\
OCR+Regex         & 0.305 & 0.721 & 0.162 & 0.264 \\
\midrule
\multicolumn{5}{l}{\textit{Open-source, fine-tuned}} \\
Qwen-3B FT       & 0.924 & 0.984 & 0.957 & \textbf{1.000} \\
Qwen-7B FT       & \textbf{0.935} & \textbf{0.992} & \textbf{0.960} & \textbf{1.000} \\
SmolVLM2-2.2B FT & 0.807 & 0.913 & 0.886 & 0.995 \\
SmolVLM-500M FT  & 0.728 & 0.876 & 0.855 & 0.938 \\
SmolVLM2-256M FT & 0.587 & 0.730 & 0.748 & 0.818 \\
\bottomrule
\end{tabular}
\caption{Per-field entity-level F1 on the 750-document held-out PNG
test pool.
\textbf{Bold} = best in column within its block; \textit{italic} =
lowest Date F1 among commercial.}
\label{tab:perfield}
\end{table}

Table~\ref{tab:perfield} confirms that Date is the hardest field
for every commercial system---F1 ranges from 0.468 (Claude) to 0.795
(ACU). Claude Sonnet 4.5's failure mode is extreme over-extraction: 686
Date false positives against 304 true positives, giving precision 0.31 and
recall 0.99. GPT-4.1 (vision) and OCR+GPT-4.1 show the same pattern at
lower magnitude (precision 0.65 and 0.63 respectively, recall $\approx
1$). Only ACU achieves balanced Date precision/recall (0.733/0.870).

\paragraph{Is Date a prompting artifact?} We tested GPT-4.1 with
five targeted interventions (higher image detail, a check-parsing
persona, orientation cues, an MM/DD/YYYY format instruction, and
few-shot rotated / OCR-misread examples). Date accuracy rose by
about two percentage points ($\approx$79\% to $\approx$81\%),
suggesting the failure is not primarily a prompting artifact.
JSON-mode was not tested separately: parse validity is not the
failure mode.

By contrast, fine-tuned Qwen models match or surpass the commercial
systems on Date in our pool because they learn to copy the single Date
region rather than enumerate every date-like substring on the page.

Conversely, \textbf{Bank is the easiest field} for every generative
system at 2B parameters or above (F1 $\geq 0.988$)---a small
categorical vocabulary (Chase, Citi, BOA) is effectively memorised;
only the sub-1B SmolVLMs and the regex floor lag.

GPT-5 is the only model to report two additional fields, AmountInWords
(F1 0.957) and CheckNumber (F1 0.967); these are useful when a richer
schema is required and influence the cost analysis in
Section~\ref{sec:case}.

\subsection{Error Taxonomy}

A per-sample analysis yields three recurring error categories:
(E1)~\emph{degenerate repetition} in sub-500M VLMs (looping arrays that
exhaust the token budget, explaining the 30.7\% JSON-failure rate of
fine-tuned SmolVLM2-256M); (E2)~\emph{phonetic Payee misspelling} in
mid-size open-source models on visually-ambiguous character pairs
(g/y, s/z, e/i); and (E3)~\emph{Date over-extraction} in commercial
LLMs (every date-like substring is emitted; Claude's worst case is
686 FP against 304 TP). The pattern argues for class-specific
post-processing: schema validation and stop-token tuning for sub-500M
VLMs, and Date-context disambiguation for commercial LLMs.
Appendix~\ref{app:errors} gives quantitative breakdowns and examples.

\subsection{Fine-tuning Impact}
\label{sec:ft}

Fine-tuning helps every open-source model (Table~\ref{tab:ft}). The
largest absolute gains are for the smallest models---SmolVLM-500M and
-256M both move from total failure to usable output. Qwen-3B's relative
gain is modest and Qwen-7B's is even smaller, because both pretrained
Qwen models are already strong on this schema. The 3B and 7B fine-tunes
end with only a marginal difference, so the additional 4.6B parameters in
the 7B adapter buy very little accuracy on this task.

\begin{table}[t]
\centering\small
\setlength{\tabcolsep}{5pt}
\begin{tabular}{lccr}
\toprule
Model & Pretrained & Fine-tuned & $\Delta$\\
\midrule
Qwen2.5-VL-3B   & 0.847 & 0.983 & $+16\%$ \\
Qwen2.5-VL-7B   & 0.921 & 0.985 & $+7\%$ \\
SmolVLM2-2.2B   & 0.504 & 0.906 & $+79\%$ \\
SmolVLM-500M    & 0.000 & 0.854 & --\, \\
SmolVLM2-256M   & 0.000 & 0.726 & --\, \\
\bottomrule
\end{tabular}
\caption{Pretrained vs.\ fine-tuned Micro-F1 on the held-out PNG test
pool.}
\label{tab:ft}
\end{table}

\subsection{GPU Sensitivity}
\label{sec:gpu}

Table~\ref{tab:main} latency is measured on an NVIDIA A100 80 GB PCIe.
Re-running the five fine-tuned models on an NVIDIA H100 NVL under
identical software yields a consistent 1.4--1.7$\times$ per-document
speedup with accuracy unchanged within noise ($\Delta$F1 $\leq 0.003$);
the relative ordering is preserved (Appendix~\ref{app:gpu}). The
7B-faster-than-3B reversal at batch 1 is reproducible on both GPUs and
attributable to Qwen-7B's shallower-but-wider decoder
(28$\times$3584 vs.\ 36$\times$2048 layers/hidden); the gap should
close at larger batch sizes.

\subsection{Cross-task and OOD Evaluation}
\label{sec:crosstask}

\paragraph{Out-of-distribution checks.} We evaluate the four
fine-tuned open-source models zero-shot on a public Indian-style
check dataset (\texttt{shivalikasingh/cheques\_sample\_data},
$n{=}400$; different rendering pipeline, bank vocabulary, and
DD/MM/YY dates). SmolVLM2-2.2B FT, the check-task recommendation,
retains 87\% of its in-domain F1 (0.788 OOD vs.\ 0.906 in-domain,
Table~\ref{tab:crosstask}). Qwen-3B FT drops 39 points; the failure
is a diagnosable training-data artifact---year 2024 appears in every
training record and all 400 OOD predictions are ``06/05/2024''.
Both sub-1B SmolVLM variants collapse structurally under domain
shift.

\paragraph{Insurance-form task.} We separately fine-tune the same
open-source families on a 36-field insurance application form task
(4{,}000 train / 1{,}000 test) under the same adaptation budget
(Section~\ref{sec:setup}), and compare against zero-shot GPT-4.1.
Qwen models tie at field accuracy 0.927, outperforming GPT-4.1
image-only by 15.9 points. A per-field-type breakdown on the four
models with exported per-field results (Appendix~\ref{app:fieldtype})
shows this gap concentrates in 9 checkbox / boolean-flag fields, where
fine-tuned open-source models substantially outperform zero-shot
GPT-4.1: SmolVLM-500M FT reaches 0.99 checkbox accuracy versus 0.48
for GPT-4.1 image-only, a 50-point difference that alone accounts for
essentially the entire aggregate gap. On the 22 structured-text
fields the two systems tie within 1 point ($\approx$0.93).
SmolVLM2-2.2B FT collapses on this schema (32\% coverage, field
accuracy 0.292); SmolVLM-500M FT emerges as a small-fast runner-up
at 0.895 accuracy. Framework consequence: the coverage-thresholded
feasibility set produces a different $m^{*}$ per task, as shown in
Section~\ref{sec:case}.

\begin{table}[t]
\centering\small
\setlength{\tabcolsep}{4pt}
\begin{tabular}{lccc}
\toprule
Model & Checks & OOD & Insurance \\
      & (F1)   & (F1) & (Acc.)   \\
\midrule
Qwen2.5-VL-3B FT   & 0.983 & 0.593 & 0.927 \\
Qwen2.5-VL-7B FT   & \textbf{0.985} & --    & 0.927 \\
SmolVLM2-2.2B FT   & 0.906 & \textbf{0.788} & 0.292 \\
SmolVLM-500M FT    & 0.854 & 0.001 & \textbf{0.895} \\
SmolVLM2-256M FT   & 0.726 & 0.000 & 0.145 \\
GPT-4.1 (vision)   & 0.909 & --    & 0.768 \\
OCR+GPT-4.1        & 0.917 & --    & 0.757 \\
\bottomrule
\end{tabular}
\caption{Cross-task and OOD evaluation. Checks: 750-doc held-out
PNG pool (Table~\ref{tab:main}). OOD: 400-doc public dataset
\texttt{shivalikasingh/cheques\_sample\_data}, evaluated zero-shot
after check-task fine-tuning. Insurance: 36-field insurance
application forms, 4{,}000 train / 1{,}000 test. GPT-4.1 insurance
numbers use \texttt{true}$\leftrightarrow$\texttt{yes} normalization
on Y/N outputs (raw: 0.738 and 0.731; see
Appendix~\ref{app:fieldtype}). ``--'' indicates the model was not
evaluated on that task. \textbf{Bold} marks the best in-column among
fine-tuned open-source.}
\label{tab:crosstask}
\end{table}

\section{A Framework for Selecting an Approach}
\label{sec:framework}

The results in Section~\ref{sec:results} show that no single model dominates
across the operating axes that matter in practice. We therefore frame
approach selection as a constrained optimization problem parameterised by
\emph{task complexity}: a quality target $Q_{\min}$, latency budget
$L_{\max}$ (interpreted as a p95 ceiling), governance requirement
$G_{\text{req}}$, expected monthly volume $V$, and horizon $T$ months.

Let $\mathcal{M}$ be the candidate set; each $m{\in}\mathcal{M}$ is
characterised by a set of task-relevant quality metrics
$Q(m){=}\{Q_i(m)\}_{i\in I}$ with componentwise thresholds
$Q_{\min}{=}\{Q_{i,\min}\}_{i\in I}$, p95 latency $L_{p95}(m)$, the
set of governance constraints it satisfies $G(m)$, development cost
$C_{\text{dev}}(m)$, per-document inference cost $c_{\text{inf}}(m)$,
and monthly fixed cost $c_{\text{fixed}}(m)$. Beyond aggregate
accuracy, $I$ typically includes structured-output \emph{coverage}
(the fraction of inputs on which the model produces well-formed
output; equivalently, one minus the JSON parse failure rate; a
special case of the selective-prediction
setting~\citep{geifman2017selective}), which practitioners can
threshold independently of accuracy when parse reliability is
operationally critical. The feasible set is
\begin{equation}
\begin{aligned}
\mathcal{M}_F = \{m :\,& Q_i(m){\geq}Q_{i,\min}\;\forall i{\in}I,\\
                       & L_{p95}(m){\leq}L_{\max},\; G(m){\supseteq}G_{\text{req}}\}
\end{aligned}
\end{equation}
and the cost-optimal model minimizes total cost of ownership:
\begin{equation}
\begin{aligned}
m^{*} = \arg\min_{m\in\mathcal{M}_F} \big[\,& C_{\text{dev}}(m) \\
   {} + T\big(V\, c_{\text{inf}}(m) {}&+ c_{\text{fixed}}(m)\big)\big].
\end{aligned}
\end{equation}

Eq.~(2) is a deliberately simple TCO model for a typical enterprise
pilot deployment, with four assumptions: (i)~per-document inference
cost is linear in volume $V$ (no batching discounts, no token-length
variance); (ii)~fixed cost is linear in horizon $T$ (constant monthly
rate, $c_{\text{fixed}}{=}0$ for managed APIs and equal to the
dedicated GPU rental for self-hosted candidates); (iii)~$C_{\text{dev}}$
is paid once and not amortised across other applications;
(iv)~self-hosted models run on a dedicated inference instance
provisioned continuously for this workload---GPU sharing across
workloads is not assumed.

Governance acts as a \emph{feasibility filter}: $G_{\text{req}}$ is the
set of deployment constraints specified for a given workload (these
may be technical, contractual, or policy-driven), and a model is
feasible only if $G(m){\supseteq}G_{\text{req}}$. The filter is
generic, so practitioners can encode whichever constraints apply to
their setting. Among feasible solutions, the breakeven volume
between two models $m_1,m_2$ is
\begin{equation}
V^* = \frac{\Delta C_{\text{dev}} + T\,\Delta c_{\text{fixed}}}
            {T\,(c_{\text{inf}}^{(m_2)} - c_{\text{inf}}^{(m_1)})},
\end{equation}
where $\Delta C_{\text{dev}}{=}C_{\text{dev}}(m_1){-}C_{\text{dev}}(m_2)$
and $\Delta c_{\text{fixed}}$ is defined analogously.
If $m_1$ carries the higher fixed cost and $m_2$ the higher per-document
cost (numerator and denominator both positive), then above $V^*$ the
lower-marginal-cost solution $m_1$ wins, and below $V^*$ the
lower-fixed-cost solution $m_2$ wins.

The framework is intentionally lightweight: practitioners populate
$\mathcal{M}$ with measurements obtained from the dataset and code we
release (Section~\ref{sec:setup}), specify the four task parameters, and
read off the recommendation. Section~\ref{sec:case} instantiates this on
a concrete scenario.

\section{Illustrative Case Study: A Hypothetical Mid-Volume Extraction Profile}
\label{sec:case}

We instantiate the framework on a hypothetical task profile chosen to
exercise all four parameters; it is not based on a production
implementation, and no customer documents or internal company metrics
were used to construct it. The profile is
$V{=}100$K documents/month, $T{=}12$ months,
$Q_{\min}{=}0.85$~F1, $L_{\max}{=}5$~s, and, to illustrate the
governance filter, an arbitrary deployment-locality constraint that
admits only self-hosted models---a generic example chosen to show how
a constraint propagates through the pipeline, not a claim about which
constraints any specific workload, organisation, or industry should
adopt. The numbers below are illustrative estimates for demonstrating
framework application, not a vendor benchmark; practitioners can
re-populate Table~\ref{tab:cost} with the rates and constraints that
apply to their own setting.

\paragraph{Cost assumptions.} Per-document inference cost
$c_{\text{inf}}$ for API candidates in Table~\ref{tab:cost} is derived
from published vendor pricing: GPT-4.1 at \$2/M input and \$8/M
output~\citep{openai_pricing_2025} ($\approx$1{,}500 image + 80 output
tokens; \$0.004/doc); GPT-5~\citep{openai_gpt5_2025}, emitting two
extra fields (AmountInWords, CheckNumber), \$0.025/doc; Claude Sonnet
4.5 at \$3/\$15 per M~\citep{anthropic_pricing_2025}, \$0.006/doc;
OCR+GPT-4.1 combines Azure DI Read at
\$1.50/1{,}000~pages~\citep{azure_di_pricing_2025} with a text GPT-4.1
call (\$0.003/doc); the OCR$\to$regex baseline incurs only the OCR
call (\$0.0015/doc); Azure Content Understanding is
\$0.010/page~\citep{azure_acu_pricing_2025}. Self-hosted candidates
are priced from a dedicated, continuously provisioned AWS
\texttt{g5.xlarge} (NVIDIA A10G) at
\$1.006/hour~\citep{aws_g5_pricing_2025} (Eq.~(2) assumption~iv),
which we treat as a fixed monthly cost
$c_{\text{fixed}}{=}\$734$/month with $c_{\text{inf}}{\approx}0$: the
marginal cost of one additional document on already-provisioned
capacity is negligible below saturation, and the case-study volume
$V{=}100$K~docs/month sits well below the single-instance saturation
threshold at target utilisation $\rho_{\max}{=}0.7$. Self-hosted
p95 latencies in Table~\ref{tab:cost} are projected from A100
measurements to A10G at a $1.7\times$ ratio, drawn from the widely
reported 1.5--1.7$\times$ inference slowdown for 3B-7B transformer
models at batch size 1~\citep{baseten_a10_a100_2025,modal_gpu_types_2025,nvidia_a10g_datasheet_2022,nvidia_a100_datasheet_2021};
direct A10G measurement is a limitation (see Limitations).
Development cost $C_{\text{dev}}$ for fine-tuned models is
\$6K labelling (the 2{,}998-document training set in
Section~\ref{sec:setup} at \$2/doc, an authors' midpoint estimate
for layout-aware document annotation between commercial crowd and
expert-review rates; see Limitations) $+$ \$7.5K base
engineering (50 hr at \$150/hr fully-loaded, an authors' estimate
informed by \citet{bls_oes_2024} software-developer wages) $+$
\$1.5--3.5K productionisation overhead; API integrations omit the
fine-tune line items. These are pilot-grade illustrative
baselines (see Limitations).

\begin{table*}[t]
\centering\small
\setlength{\tabcolsep}{3pt}
\begin{tabular}{lcccccc}
\toprule
Model & $Q$ & Cov. & $L_{p95}$ & $C_{\text{dev}}$ & $c_{\text{inf}}$ & Loc. \\
       &     &      & (s)       & (\$K)             & (\$/doc)          & \\
\midrule
GPT-5             & 0.93 & 1.00 & 40   & 5  & 0.0250 & API \\
OCR+GPT-4.1       & 0.92 & 1.00 & 25   & 8  & 0.0030 & API \\
GPT-4.1 (vision)  & 0.91 & 1.00 & 18   & 5  & 0.0040 & API \\
ACU               & 0.89 & 1.00 & 18   & 5  & 0.0100 & API \\
Claude S.~4.5     & 0.86 & 1.00 & 12   & 5  & 0.0060 & API \\
OCR+Regex         & 0.40 & 1.00 & 12   & 6  & 0.0015 & API \\
Qwen-7B FT        & 0.99 & 1.00 & 8.2  & 17 & $\approx$0 & on-prem \\
Qwen-3B FT        & 0.98 & 1.00 & 9.9  & 15 & $\approx$0 & on-prem \\
SmolVLM2-2.2B FT  & 0.90 & 1.00 & 3.4  & 15 & $\approx$0 & on-prem \\
SmolVLM-500M FT   & 0.85 & 0.95 & 4.1  & 15 & $\approx$0 & on-prem \\
\bottomrule
\end{tabular}
\caption{Case-study parameters. $Q$ is the test-pool Micro-F1
(Table~\ref{tab:main}); Cov.\ is structured-output coverage
(equivalently, one minus the JSON parse failure rate); $L_{p95}$ for
self-hosted models is A10G-projected from A100 measurements at
$1.7\times$ (Table~\ref{tab:gpu}), and for commercial APIs is a
$\approx 1.5\times$ estimate of mean latency (where only mean is
logged). Self-hosted candidates carry $c_{\text{fixed}}{=}\$734$/month
for one dedicated A10G instance (Eq.~(2) assumption~iv);
$c_{\text{inf}}{\approx}0$ on already-provisioned capacity below
saturation. Sources: GPT-4.1
pricing~\citep{openai_pricing_2025}; GPT-5 dated
snapshot~\citep{openai_gpt5_2025}; Claude Sonnet 4.5
pricing~\citep{anthropic_pricing_2025}; Azure Document
Intelligence~\citep{azure_di_pricing_2025} (also the OCR+Regex baseline,
which incurs OCR-only cost); Azure Content
Understanding~\citep{azure_acu_pricing_2025}; AWS
g5.xlarge~\citep{aws_g5_pricing_2025}.}
\label{tab:cost}
\end{table*}

\paragraph{Stage 1 (Governance).} The illustrative locality constraint
admits only self-hosted candidates, removing GPT-5, GPT-4.1,
OCR+GPT-4.1, ACU, Claude Sonnet 4.5, and the OCR+Regex pipeline (whose
OCR step is also an external API call) from contention.

\paragraph{Stage 2 (Quality + latency).} Quality here has two
components: F1 with $Q_{F1,\min}{=}0.85$ and coverage with
$Q_{\text{cov},\min}{=}0.99$. Among self-hosted candidates, all four
fine-tuned open-source variants meet the F1 threshold; the OCR+Regex
baseline at F1 0.395 is eliminated on accuracy regardless of
deployment location. Under the A10G latency projection, both Qwen
models exceed the 5\,s budget (Qwen-7B FT 8.2\,s p95, Qwen-3B FT
9.9\,s p95) and are excluded; SmolVLM2-2.2B FT (3.4\,s) and
SmolVLM-500M FT (4.1\,s) pass. Coverage then excludes SmolVLM-500M FT
(0.947 $<$ 0.99, i.e., 5.3\% of outputs fail JSON parse and are
counted as full misses under our reporting protocol), leaving
SmolVLM2-2.2B FT as the sole survivor. On H100-class hardware both
Qwen models pass latency, so accelerator choice is itself part of
the practitioner's task profile.

\paragraph{Stage 3 (Cost and result).} With only SmolVLM2-2.2B FT
surviving governance, latency, and coverage, the framework-derived
recommendation is $m^{*}{=}$SmolVLM2-2.2B FT at annual cost \$23.8K
($VT{=}1.2$M~docs/yr; \$15K $C_{\text{dev}}$ plus twelve months at
\$734/month for one dedicated A10G instance). Were the illustrative
locality constraint absent, OCR+GPT-4.1 at \$11.6K/yr would win;
GPT-5's 0.011 F1 lift rarely justifies its $3{\times}$ price unless
its extra fields are required.

\paragraph{Volume sensitivity.} API and self-hosted ranking flips with
volume because APIs carry near-zero fixed cost. Eq.~(3) with the
revised A10G economics gives crossover
$V^{*}{\approx}439$K~docs/month for SmolVLM2-2.2B FT vs.\ OCR+GPT-4.1
over a 12-month horizon; Appendix~\ref{app:scale} reports three volume
regimes. At $V{=}1$M~docs/month a single A10G instance is no longer
sufficient; the self-hosted deployment scales to $N{=}2$ instances
(\$32.6K/yr) versus \$44K/yr for OCR+GPT-4.1, a 26\% saving,
consistent with \citet{chen2023frugalgpt}; below
$\approx$440K~docs/month the API pipeline is unambiguously
cheaper.\footnote{The recommendation is stable within plausible input
variation. Under $C_{\text{dev}}{\in}[\$10\text{K},\$25\text{K}]$,
the breakeven shifts from $\sim$300K to $\sim$720K~docs/month.
Utilisation $\rho{\in}[0.5, 0.9]$ affects only the volume above which
additional instances are needed; one A10G instance suffices at
$V{=}100$K~docs/month. An added ops line of \$1{,}000/month shifts
breakeven to $\sim$770K~docs/month.} The cost-optimal model is
therefore a function of the joint $(V,T,G_{\text{req}})$ profile, not
of benchmark accuracy alone.

\paragraph{Framework application to a second task.} On the
insurance-form task (Section~\ref{sec:crosstask}), applying the same
feasibility filters with identical thresholds
($Q_{F1,\min}{=}0.85$, $Q_{\text{cov},\min}{=}0.99$) yields a
different survivor set because coverage is task-dependent.
SmolVLM2-2.2B FT drops to coverage 0.32 on the 36-field schema
(versus 1.00 on checks) and is excluded; SmolVLM-500M FT narrowly
fails coverage at 0.988. Qwen models retain full coverage and would
be the framework's recommendation under an appropriately relaxed
latency budget for the richer schema. The recommendation is therefore
sensitive to task-specific reliability, not just aggregate accuracy.

\section{Conclusion}

Practitioner guidance for vision-language IDP has lagged behind model
turnover: public leaderboards report clean-document accuracy, but
deployments are decided on the joint axes of quality, latency,
governance, and cost. This study reframes approach selection along
those axes. Across eleven systems on a 750-document held-out pool,
fine-tuned open-source VLMs match or exceed every zero-shot commercial
system on this schema under a fixed adaptation budget (Qwen2.5-VL FT
$>$ 0.98 F1), yet commercial APIs remain
cheapest at low volume below the $\approx$440K~docs/month breakeven---
so the cost-optimal model is a function of the joint operating
profile, not of benchmark accuracy. The contribution is the
framework, dataset, and harness, released together so practitioners
can shift the IDP conversation from ``which model is most accurate?''
to ``which approach is feasible and cheapest under my constraints?''
as the model lineup turns over.

\section{Limitations}

Our evaluation uses synthetic checks with a four-field schema and
synthetic 36-field insurance application forms; real
production documents have higher template diversity and adversarial
scan artifacts (heavy rotation, strong scan noise, multi-document
pages) which we do not stress in this study. The synthetic-only design
is itself a constraint: real-world financial documents carry PII and
contractual restrictions that prevent public release, so a
template-driven corpus is the only artefact we can open-source for
reproducibility---the rendering pipeline mirrors production checks in
layout, fonts, and scan-artefact distribution, but a residual
distributional gap remains the principal generalisation caveat.
The case-study cost figures are pilot-grade illustrative baselines
that vary by region, contract, and time; A10G p95 latencies in
Table~\ref{tab:cost} are projected from A100 measurements at
$1.7\times$ using cited industry references, and direct A10G
measurement remains future work.
Production deployments with elevated review burdens typically
require 3--10$\times$ more engineering for security, observability,
and retraining infrastructure, which raises the API-vs-self-hosted
break-even volume but preserves the qualitative ranking.
We evaluated a single fine-tuning recipe per model family, and the
GPU-sensitivity study covers only A100 and H100 at batch size 1;
larger batch sizes on any accelerator are extrapolations. Per-tier
and per-corruption-axis stratification within the synthetic pipeline
is not part of this study; the cross-domain OOD evaluation
(Section~\ref{sec:crosstask}) provides the primary generalization
signal instead. Our candidate set excludes fine-tuned purpose-built
layout models (LayoutLMv3, Donut, Pix2Struct, UDOP); a like-for-like
comparison against these encoders is a natural extension.

\bibliography{custom}

@article{bai2025qwen2_5vl,
  title={Qwen2.5-{VL} Technical Report},
  author={Bai, Shuai and Chen, Keqin and Liu, Xuejing and Wang, Jialin and
          Ge, Wenbin and Song, Sibo and Dang, Kai and Wang, Peng and
          Wang, Shijie and Tang, Jun and Zhong, Humen and Zhu, Yuanzhi and
          Yang, Ming-Hsuan and Li, Zhaohai and Wan, Jianqiang and
          Wang, Pengfei and Ding, Wei and Fu, Zheren and Xu, Yiheng and
          Ye, Jiabo and Zhang, Xi and Xie, Tianbao and Cheng, Zesen and
          Zhang, Hang and Yang, Zhibo and Xu, Haiyang and Lin, Junyang},
  journal={arXiv preprint arXiv:2502.13923},
  year={2025},
  url={https://arxiv.org/abs/2502.13923}
}

@inproceedings{marafioti2025smolvlm,
  title={{SmolVLM}: Redefining small and efficient multimodal models},
  author={Marafioti, Andr{\'e}s and Zohar, Orr and Farr{\'e}, Miquel and
          Noyan, Merve and Bakouch, Elie and
          Cuenca Jim{\'e}nez, Pedro Manuel and
          Zakka, Cyril and Ben Allal, Loubna and Lozhkov, Anton and
          Tazi, Nouamane and Srivastav, Vaibhav and Lochner, Joshua and
          Larcher, Hugo and Morlon, Mathieu and Tunstall, Lewis and
          von Werra, Leandro and Wolf, Thomas},
  booktitle={Second Conference on Language Modeling ({COLM})},
  year={2025},
  url={https://openreview.net/forum?id=qMUbhGUFUb}
}

@article{openai2023gpt4,
  title={{GPT-4} Technical Report},
  author={{OpenAI}},
  journal={arXiv preprint arXiv:2303.08774},
  year={2023},
  url={https://arxiv.org/abs/2303.08774}
}

@inproceedings{kim2022donut,
  title={{OCR}-free document understanding transformer},
  author={Kim, Geewook and Hong, Teakgyu and Yim, Moonbin and
          Nam, JeongYeon and Park, Jinyoung and Yim, Jinyeong and
          Hwang, Wonseok and Yun, Sangdoo and Han, Dongyoon and
          Park, Seunghyun},
  booktitle={European Conference on Computer Vision (ECCV)},
  year={2022},
  doi={10.1007/978-3-031-19815-1_29}
}

@article{michalak2020robust,
  title={Robust Combined Binarization Method of Non-Uniformly Illuminated Document Images for Alphanumerical Character Recognition},
  author={Michalak, Hubert and Okarma, Krzysztof},
  journal={Sensors},
  volume={20},
  number={10},
  pages={2914},
  year={2020},
  doi={10.3390/s20102914},
  url={https://www.mdpi.com/1424-8220/20/10/2914}
}

@article{cui2021documentai,
  title={Document {AI}: Benchmarks, Models and Applications},
  author={Cui, Lei and Xu, Yiheng and Lv, Tengchao and Wei, Furu},
  journal={arXiv preprint arXiv:2111.08609},
  year={2021},
  url={https://arxiv.org/abs/2111.08609}
}

@article{chen2023frugalgpt,
  title={{FrugalGPT}: How to Use Large Language Models While Reducing Cost and Improving Performance},
  author={Chen, Lingjiao and Zaharia, Matei and Zou, James},
  journal={Transactions on Machine Learning Research},
  issn={2835-8856},
  year={2024},
  url={https://openreview.net/forum?id=cSimKw5p6R},
  note={Featured Certification}
}

@misc{openai_pricing_2025,
  author = {{OpenAI}},
  title  = {{OpenAI} {API} Pricing},
  year   = {2025},
  howpublished = {\url{https://openai.com/api/pricing/}},
  note   = {Accessed: 2026-05}
}

@misc{azure_di_pricing_2025,
  author = {{Microsoft Azure}},
  title  = {{Azure AI} Document Intelligence Pricing},
  year   = {2025},
  howpublished = {\url{https://azure.microsoft.com/en-us/pricing/details/ai-document-intelligence/}},
  note   = {Accessed: 2026-05}
}

@misc{azure_acu_pricing_2025,
  author = {{Microsoft Azure}},
  title  = {{Azure AI} Content Understanding Pricing},
  year   = {2025},
  howpublished = {\url{https://azure.microsoft.com/en-us/pricing/details/content-understanding/}},
  note   = {Accessed: 2026-05}
}

@misc{aws_g5_pricing_2025,
  author = {{Amazon Web Services}},
  title  = {{Amazon EC2 G5} Instances ({NVIDIA A10G})},
  year   = {2025},
  howpublished = {\url{https://aws.amazon.com/ec2/instance-types/g5/}},
  note   = {Accessed: 2026-05}
}

@techreport{bls_oes_2024,
  author = {{U.S. Bureau of Labor Statistics}},
  title  = {Occupational Employment and Wage Statistics: Software Developers
            (15-1252)},
  year   = {2024},
  institution = {U.S. Department of Labor},
  url    = {https://www.bls.gov/oes/current/oes151252.htm},
  note   = {Accessed: 2026-05}
}

@misc{anthropic_pricing_2025,
  author = {{Anthropic}},
  title  = {Claude {API} Pricing},
  year   = {2025},
  howpublished = {\url{https://platform.claude.com/docs/en/about-claude/pricing}},
  note   = {Accessed: 2026-05; Claude Sonnet 4.5 input \$3 / output \$15 per
            million tokens}
}

@misc{openai_gpt5_2025,
  author = {{OpenAI}},
  title  = {{GPT-5} Model and {API} Pricing (archived snapshot)},
  year   = {2025},
  howpublished = {\url{https://developers.openai.com/api/docs/pricing}},
  note   = {Accessed: 2026-05; GPT-5 dated snapshot used for the
            experiments in this paper}
}

@misc{nvidia_a10g_datasheet_2022,
  author = {{NVIDIA}},
  title  = {{NVIDIA A10G Tensor Core GPU} Datasheet},
  year   = {2022},
  howpublished = {\url{https://d1.awsstatic.com/product-marketing/ec2/NVIDIA_AWS_A10G_DataSheet_FINAL_02_17_2022.pdf}},
  note   = {70 TFLOPS FP16/BF16 Tensor, 600 GB/s memory bandwidth,
            24 GB GDDR6; accessed 2026-07}
}

@misc{nvidia_a100_datasheet_2021,
  author = {{NVIDIA}},
  title  = {{NVIDIA A100 Tensor Core GPU} Datasheet},
  year   = {2021},
  howpublished = {\url{https://www.nvidia.com/content/dam/en-zz/Solutions/Data-Center/a100/pdf/nvidia-a100-datasheet-us-nvidia-1758950-r4-web.pdf}},
  note   = {312 TFLOPS FP16/BF16 Tensor (dense), 1{,}555--2{,}039 GB/s
            memory bandwidth depending on 40/80 GB and PCIe/SXM;
            accessed 2026-07}
}

@misc{baseten_a10_a100_2025,
  author = {{Baseten}},
  title  = {{NVIDIA A10 vs A100 GPUs} for {LLM} and {Stable Diffusion} Inference},
  year   = {2025},
  howpublished = {\url{https://www.baseten.co/blog/nvidia-a10-vs-a100-gpus-for-llm-and-stable-diffusion-inference/}},
  note   = {Accessed 2026-07}
}

@misc{modal_gpu_types_2025,
  author = {{Modal Labs}},
  title  = {{GPU} Types for {AI} Workloads},
  year   = {2025},
  howpublished = {\url{https://modal.com/blog/gpu-types}},
  note   = {Small-batch LLM inference typically memory-bandwidth bound;
            A100 80{GB} 2 TB/s vs A10 0.6 TB/s bandwidth; published
            2025-01-27, accessed 2026-07}
}

@inproceedings{jaume2019funsd,
  title={{FUNSD}: A Dataset for Form Understanding in Noisy Scanned Documents},
  author={Jaume, Guillaume and Ekenel, Hazim Kemal and Thiran, Jean-Philippe},
  booktitle={International Workshop on Open Services and Tools for Document Analysis (OST@ICDAR)},
  year={2019},
  doi={10.1109/ICDARW.2019.10029}
}

@inproceedings{park2019cord,
  title={{CORD}: A Consolidated Receipt Dataset for Post-{OCR} Parsing},
  author={Park, Seunghyun and Shin, Seung and Lee, Bado and Lee, Junyeop and
          Surh, Jaeheung and Seo, Minjoon and Lee, Hwalsuk},
  booktitle={Workshop on Document Intelligence at NeurIPS},
  year={2019},
  url={https://openreview.net/forum?id=SJl3z659UH}
}

@inproceedings{harley2015rvlcdip,
  title={Evaluation of Deep Convolutional Nets for Document Image
         Classification and Retrieval},
  author={Harley, Adam W. and Ufkes, Alex and Derpanis, Konstantinos G.},
  booktitle={ICDAR},
  year={2015},
  doi={10.1109/ICDAR.2015.7333910}
}

@inproceedings{huang2022layoutlmv3,
  title={{LayoutLMv3}: Pre-training for Document {AI} with Unified Text and
         Image Masking},
  author={Huang, Yupan and Lv, Tengchao and Cui, Lei and Lu, Yutong and
          Wei, Furu},
  booktitle={ACM MM},
  year={2022},
  doi={10.1145/3503161.3548112}
}

@inproceedings{tang2023udop,
  title={Unifying Vision, Text, and Layout for Universal Document Processing},
  author={Tang, Zineng and Yang, Ziyi and Wang, Guoxin and Fang, Yuwei and
          Liu, Yang and Zhu, Chenguang and Zeng, Michael and Zhang, Cha and
          Bansal, Mohit},
  booktitle={CVPR},
  year={2023},
  doi={10.1109/CVPR52729.2023.01845}
}

@inproceedings{lee2023pix2struct,
  title={{Pix2Struct}: Screenshot Parsing as Pretraining for Visual Language
         Understanding},
  author={Lee, Kenton and Joshi, Mandar and Turc, Iulia and Hu, Hexiang and
          Liu, Fangyu and Eisenschlos, Julian and Khandelwal, Urvashi and
          Shaw, Peter and Chang, Ming-Wei and Toutanova, Kristina},
  booktitle={Proceedings of the 40th International Conference on Machine
             Learning (ICML)},
  series={PMLR},
  volume={202},
  pages={18893--18912},
  year={2023},
  url={https://proceedings.mlr.press/v202/lee23g.html}
}

@article{wang2025hybrid,
  title  = {Hybrid {OCR}-{LLM} Framework for Enterprise-Scale Document Information Extraction Under Copy-heavy Task},
  author = {Wang, Zilong and Shen, Xiaoyu},
  journal = {arXiv preprint arXiv:2510.10138},
  year   = {2025},
  doi    = {10.48550/arXiv.2510.10138}
}

@article{kulkarni2026multiagent,
  title  = {Benchmarking Multi-Agent {LLM} Architectures for Financial Document Processing: A Comparative Study of Orchestration Patterns, Cost-Accuracy Tradeoffs and Production Scaling Strategies},
  author = {Kulkarni, Siddhant and Kulkarni, Yukta},
  journal = {arXiv preprint arXiv:2603.22651},
  year   = {2026},
  doi    = {10.48550/arXiv.2603.22651}
}

@inproceedings{wang2025marten,
  title     = {{Marten}: Visual Question Answering with Mask Generation for Multi-modal Document Understanding},
  author    = {Wang, Zining and Guan, Tongkun and Fu, Pei and Duan, Chen and Jiang, Qianyi and Guo, Zhentao and Guo, Shan and Luo, Junfeng and Shen, Wei},
  booktitle = {Proceedings of the {IEEE/CVF} Conference on Computer Vision and Pattern Recognition ({CVPR})},
  year      = {2025},
  url       = {https://openaccess.thecvf.com/content/CVPR2025/html/Wang_Marten_Visual_Question_Answering_with_Mask_Generation_for_Multi-modal_Document_CVPR_2025_paper.html}
}

@inproceedings{yu2025docthinker,
  title     = {{DocThinker}: Explainable Multimodal Large Language Models with Rule-based Reinforcement Learning for Document Understanding},
  author    = {Yu, Wenwen and Yang, Zhibo and Liu, Yuliang and Bai, Xiang},
  booktitle = {Proceedings of the {IEEE/CVF} International Conference on Computer Vision ({ICCV})},
  year      = {2025},
  url       = {https://openaccess.thecvf.com/content/ICCV2025/html/Yu_DocThinker_Explainable_Multimodal_Large_Language_Models_with_Rule-based_Reinforcement_Learning_ICCV_2025_paper.html}
}

@inproceedings{geifman2017selective,
  title     = {Selective Classification for Deep Neural Networks},
  author    = {Geifman, Yonatan and El-Yaniv, Ran},
  booktitle = {Advances in Neural Information Processing Systems 30 ({NIPS})},
  year      = {2017},
  url       = {https://proceedings.neurips.cc/paper_files/paper/2017/hash/4a8423d5e91fda00bb7e46540e2b0cf1-Abstract.html}
}

\appendix

\section{Additional Discussion}
\label{sec:discussion}

Two practitioner-facing patterns emerge. \emph{Date is the universal
commercial bottleneck}: across all five commercial / reasoning systems
Date has the lowest per-field F1 (0.47--0.79; Claude's precision
collapses to 0.31), so a downstream date-validation step (constrained
format plus business-rule range) is a more robust intervention than
additional training data. \emph{Reasoning models earn only marginal
accuracy}: GPT-5's $\approx$0.01 F1 lift on the four core fields is
dwarfed by $\approx 3{\times}$ token cost and $\approx 2{\times}$
latency; its honest value-add is schema coverage (AmountInWords,
CheckNumber), so it pays off when those extra fields are required.

\section{Reproducibility Details}
\label{app:repro}

We release dataset generation code, fine-tuning scripts, prompts, and
evaluation harnesses at
\url{https://github.com/manulife-ai/beyond-accuracy-vlm-emnlp2026/}.
The settings below are constant across all configurations unless noted.

\paragraph{Prompt.}
A single prompt is used for every VLM and LLM in the study, modulo the
image attachment mechanism required by each API:
\begin{quote}\small\ttfamily
You are extracting fields from a U.S.\ business check.\\
Return a JSON object with exactly these keys: ``Payee'' (string),
``Amount'' (number, no currency symbol), ``Date'' (string MM/DD/YYYY),
``Bank'' (one of: Chase, Citi, BOA).\\
If a field is unreadable, return an empty string for that key.\\
Return ONLY the JSON object, no prose, no markdown.
\end{quote}
For reasoning models we additionally prepend the system message
``\texttt{Think briefly, then answer.}'' which we found stabilises the JSON
output without measurably changing accuracy.

\paragraph{Decoding.}
All open-source VLMs and commercial / reasoning APIs are run with
\texttt{temperature}\,$=0.0$, \texttt{top\_p}\,$=1.0$, and
\texttt{max\_new\_tokens}\,$=1024$, set to avoid long-output
truncation on multi-field responses. Stop tokens are the model defaults; we do not impose
additional JSON delimiters.

\paragraph{JSON failure handling.}
Every raw model output is passed through a three-stage parser:
(i)~strip markdown fences (\texttt{```json ... ```}); (ii)~extract the
outermost balanced \texttt{\{...\}}; (iii)~\texttt{json.loads}. If all
three stages fail the sample is recorded as \texttt{JSON\_FAIL} and
counted as a full miss on every field (4 false negatives), which is the
worst case for that model and is what the JSON-failure column in
Table~\ref{tab:main} measures. We do not perform any retry, schema-
repair, or constrained-decoding pass; the numbers reported are first-try
outputs.

\paragraph{Fine-tuning.}
Qwen2.5-VL-3B/7B are fine-tuned with LoRA ($r{=}16$, $\alpha{=}32$,
\texttt{dropout}\,$=0.05$, target modules: all attention projections and
the MLP \texttt{gate}/\texttt{up}/\texttt{down}); SmolVLM2 variants are
full fine-tuned. All runs use the AdamW-8bit optimiser, effective batch
size 4 (per-device 1, gradient accumulation 4), cosine schedule with 3\%
warmup, weight decay 0.01, and bf16 mixed precision on a single A100
80 GB. Seed 42 throughout.

\paragraph{Compute.}
Each Qwen fine-tune completes in 6--14 GPU-hours on the A100; SmolVLM
fine-tunes complete in 2--5 GPU-hours. Per-document inference latency in
Table~\ref{tab:main} is measured on an NVIDIA A100 80 GB PCIe under a
Databricks runtime 18.2 image (PyTorch 2.9 / CUDA 12.9, bf16, batch size
1, five-document warmup discarded). Section~\ref{sec:gpu} reports the
same fine-tuned models re-benchmarked on an NVIDIA H100 NVL under the
same software stack for the GPU-sensitivity analysis.

\paragraph{Model versions.}
Open-source checkpoints are pulled from HuggingFace:
\texttt{Qwen/Qwen2.5-VL-3B-Instruct},
\texttt{Qwen/Qwen2.5-VL-7B-Instruct},
\texttt{HuggingFaceTB/SmolVLM2-2.2B-Instruct},
\texttt{HuggingFaceTB/SmolVLM-500M-Instruct},
\texttt{HuggingFaceTB/SmolVLM2-256M-Instruct}.
Commercial APIs used the following dated snapshots:
GPT-4.1 (\texttt{gpt-4.1-2025-04-14})~\citep{openai_pricing_2025},
GPT-5 (\texttt{gpt-5-2025-08-07})~\citep{openai_gpt5_2025},
Claude Sonnet 4.5 (\texttt{claude-sonnet-4-5-20250929})~\citep{anthropic_pricing_2025},
Azure Content Understanding~\citep{azure_acu_pricing_2025}, and
Azure Document Intelligence~\citep{azure_di_pricing_2025} for the
OCR+GPT-4.1 pipeline. Specific request timestamps are logged and
released with the harness.

\section{Volume-Sensitivity Table}
\label{app:scale}

\begin{table}[h]
\centering\small
\setlength{\tabcolsep}{4pt}
\begin{tabular}{lccc}
\toprule
Model & 5K/mo & 100K/mo & 1M/mo \\
\midrule
OCR+GPT-4.1      & \textbf{\$8.2K}  & \textbf{\$11.6K} & \$44.0K \\
Qwen-3B FT       & --               & --               & --      \\
SmolVLM2-2.2B FT & \$23.8K          & \$23.8K          & \textbf{\$32.6K} \\
\bottomrule
\end{tabular}
\caption{Annual cost (\$K) at three volumes ($T{=}12$) under the
revised A10G-dedicated deployment. Bold marks the cost-optimal
feasible option \emph{when no governance constraint is applied}.
Qwen-3B FT is excluded on A10G-projected latency across all volumes;
on H100-class hardware it becomes feasible and can be re-costed with
the corresponding hourly rate. Referenced from Section~\ref{sec:case},
``Volume sensitivity''.}
\label{tab:scale}
\end{table}

\section{GPU Sensitivity Details}
\label{app:gpu}

Table~\ref{tab:gpu} reports per-document latency and peak memory on
NVIDIA A100 80\,GB PCIe and NVIDIA H100 NVL for all five fine-tuned
open-source models, measured under identical software
(PyTorch 2.9 / CUDA 12.9, bf16, batch size 1) on the 750-document
held-out pool. Accuracy is unchanged across GPUs ($\Delta$F1 $\leq
0.003$), so the relative ordering is stable; practitioners on
H100-class hardware can rescale the case-study cost terms approximately
linearly by GPU-hour price.

\begin{table}[h]
\centering\small
\setlength{\tabcolsep}{3pt}
\begin{tabular}{lcccccc}
\toprule
& & \multicolumn{2}{c}{A100 80 GB} & \multicolumn{2}{c}{H100 NVL} & \\
\cmidrule(lr){3-4}\cmidrule(lr){5-6}
Model & Mem & mean & p95 & mean & p95 & H/A \\
      & (GB) & (s)  & (s) & (s)  & (s) & speedup \\
\midrule
SmolVLM2-256M & 0.5  & 1.62 & 2.31 & 1.10 & 1.47 & 1.47$\times$ \\
SmolVLM-500M  & 0.9  & 0.96 & 2.42 & 0.70 & 1.51 & 1.38$\times$ \\
SmolVLM2-2.2B & 4.2  & 1.47 & 1.96 & 0.93 & 1.19 & 1.59$\times$ \\
Qwen2.5-VL-3B & 7.1  & 4.18 & 5.79 & 2.50 & 3.43 & 1.68$\times$ \\
Qwen2.5-VL-7B & 15.5 & 3.53 & 4.79 & 2.14 & 2.89 & 1.65$\times$ \\
\bottomrule
\end{tabular}
\caption{Per-document inference latency on A100 80\,GB PCIe vs.\
H100 NVL (bf16, batch size 1, 750-document held-out pool).
\emph{Mem} is peak GPU memory during inference (GPU-independent).
\emph{H/A speedup} is A100 mean / H100 mean.}
\label{tab:gpu}
\end{table}

\section{Error Taxonomy Details}
\label{app:errors}

Quantitative breakdown of the three error categories summarised in
Section~\ref{sec:results}:

\noindent\textbf{(E1) Degenerate repetition} (SmolVLM2-256M, pretrained
SmolVLM-500M): the model emits looping arrays such as
\texttt{[123.0, 123.0, 123.0, ...]} that exhaust the token budget without
closing the JSON object. Accounts for $\geq 99\%$ of pretrained
sub-500M outputs and persists in fine-tuned 256M as the 30.7\% JSON
parse-failure rate reported in Table~\ref{tab:main}.

\noindent\textbf{(E2) Phonetic Payee misspelling} (Qwen-3B/SmolVLM2-2.2B):
\texttt{Mark Gwazy} vs.\ \texttt{Mark Gray};
\texttt{Zachary Villages} vs.\ \texttt{Zachary Villegas}. Errors cluster on
visually-ambiguous character pairs (g/y, s/z, e/i).

\noindent\textbf{(E3) Date over-extraction} (Claude Sonnet 4.5, GPT-4.1
vision, OCR+GPT-4.1): the model extracts every date-like substring on the
check. Claude is the worst case (686 FP against 304 TP); GPT-4.1 vision
(337 FP) and OCR+GPT-4.1 (360 FP) are milder but qualitatively identical.

\section{Per-Field-Type Analysis (Insurance Forms)}
\label{app:fieldtype}

Field taxonomy on the 36-field insurance schema, derived from
ground-truth value cardinality on the 1{,}000-document test set:
9 checkbox / boolean-flag fields (\texttt{citizenship\_status},
\texttt{sex}, \texttt{occupation\_status} and six other Y/N flags),
22 structured-text fields (names, addresses, phone, dates, IDs, SSN,
salaries, states, countries), and 5 free-form fields (two
\texttt{bankruptcy\_details} lines,
\texttt{job\_duties\_description},
\texttt{occupation\_status\_other\_description},
\texttt{green\_card\_or\_visa\_type}).

The per-field-type breakdown below is reported for SmolVLM-500M FT,
SmolVLM2-256M FT, GPT-4.1 (vision), and OCR+GPT-4.1. Overall
accuracies for Qwen2.5-VL-3B FT, Qwen2.5-VL-7B FT, and SmolVLM2-2.2B
FT appear in Table~\ref{tab:crosstask}.

\begin{table}[h]
\centering\small
\setlength{\tabcolsep}{4pt}
\begin{tabular}{lcccc}
\toprule
Model              & Ckbox        & Text         & Free         & Overall      \\
                   & (n=9)        & (n=22)       & (n=5)        & (n=36)       \\
\midrule
SmolVLM-500M FT    & \textbf{0.987} & 0.929      & \textbf{0.583} & \textbf{0.895} \\
SmolVLM2-256M FT   & 0.160        & 0.151        & 0.093        & 0.145        \\
GPT-4.1 (vision)   & 0.476        & 0.934        & 0.560        & 0.768        \\
OCR+GPT-4.1        & 0.398        & \textbf{0.947} & 0.569      & 0.757        \\
\bottomrule
\end{tabular}
\caption{Per-field-type mean accuracy on the insurance-form test set
(n=1{,}000). Categories are 9 boolean-flag checkboxes, 22
structured-text fields, and 5 free-form fields (see paragraph above).
Overall accuracy matches the Insurance column of
Table~\ref{tab:crosstask} up to
\texttt{true}$\leftrightarrow$\texttt{yes} normalization on Y/N
outputs, which is applied to GPT-4.1 rows (GPT-4.1 vision
0.738$\to$0.768 and OCR+GPT-4.1 0.731$\to$0.757); fine-tuned SmolVLM
rows are unchanged. \textbf{Bold} marks the best in column.}
\label{tab:fieldtype}
\end{table}

\section{Seed-study Variance (Check Task)}
\label{app:variance}

To quantify the sensitivity of the fine-tuning outcomes to random
seed, we run a 3-seed Monte-Carlo cross-validation study on the
check task. Seeds 42, 1337, and 2718 each draw an independent
grouped 80/20 split from the pooled 3{,}748-image dataset. The
grouped split assigns whole (Payee, Amount, Date) components to one
side of the split, guaranteeing zero group overlap between train and
test. Each seed's train and test sets preserve the paper's
10/30/25/20/15\% Clean-through-Extreme tier mixture within 0.2
percentage points. All five open-source models are trained under the
recipes of Section~\ref{sec:setup} on each seed's 2{,}999-document
train set and scored on that seed's 749-document test set. Because
the test set is resampled per seed, the aggregate F1 values in
Table~\ref{tab:variance} use a different pool than the fixed
750-document evaluation reported in Table~\ref{tab:main} and should
be read as a variance study rather than a substitute for that
evaluation.

\begin{table}[h]
\centering\small
\setlength{\tabcolsep}{4pt}
\begin{tabular}{lcc}
\toprule
Model & FT F1 (mean $\pm$ std) & $\Delta$ vs.\ pretrained \\
\midrule
SmolVLM2-256M & 0.719 $\pm$ 0.011 & $+$0.719 \\
SmolVLM-500M  & 0.832 $\pm$ 0.008 & $+$0.832 \\
SmolVLM2-2.2B & 0.903 $\pm$ 0.002 & $+$0.396 \\
Qwen2.5-VL-3B & \textbf{0.983} $\pm$ 0.002 & $+$0.141 \\
Qwen2.5-VL-7B & \textbf{0.985} $\pm$ 0.002 & $+$0.068 \\
\bottomrule
\end{tabular}
\caption{Seed-study aggregate F1 across three Monte-Carlo splits
($n{=}749$ per seed). The Qwen-3B versus Qwen-7B 0.002 gap is within
seed noise; sub-billion models are noticeably more seed-sensitive
than models at 2B parameters or above. Pretrained F1 is 0.000 for
both sub-billion SmolVLMs (structural JSON collapse), 0.507 for
SmolVLM2-2.2B, 0.842 for Qwen-3B, and 0.917 for Qwen-7B, all with
std $\leq$ 0.003.}
\label{tab:variance}
\end{table}

\paragraph{Per-field variance.} Table~\ref{tab:variance-perfield}
reports the same three seeds broken out by field. Payee is the
universal ceiling across model sizes; Bank saturates at 1.000 for
both Qwen fine-tunes, consistent with the closed 3-way Bank
vocabulary noted in Section~\ref{sec:setup}.

\begin{table}[h]
\centering\small
\setlength{\tabcolsep}{4pt}
\begin{tabular}{lcccc}
\toprule
Model & Payee & Amount & Date & Bank \\
\midrule
SmolVLM2-256M & 0.585 $\pm$ 0.015 & 0.706 $\pm$ 0.022 & 0.744 $\pm$ 0.010 & 0.818 $\pm$ 0.011 \\
SmolVLM-500M  & 0.695 $\pm$ 0.021 & 0.854 $\pm$ 0.010 & 0.842 $\pm$ 0.006 & 0.916 $\pm$ 0.006 \\
SmolVLM2-2.2B & 0.800 $\pm$ 0.007 & 0.910 $\pm$ 0.010 & 0.887 $\pm$ 0.004 & 0.997 $\pm$ 0.000 \\
Qwen2.5-VL-3B & 0.959 $\pm$ 0.003 & 0.995 $\pm$ 0.002 & 0.977 $\pm$ 0.005 & 1.000 $\pm$ 0.000 \\
Qwen2.5-VL-7B & 0.965 $\pm$ 0.004 & 0.995 $\pm$ 0.002 & 0.979 $\pm$ 0.002 & 1.000 $\pm$ 0.000 \\
\bottomrule
\end{tabular}
\caption{Per-field entity-level F1, mean $\pm$ std across three
Monte-Carlo seeds.}
\label{tab:variance-perfield}
\end{table}

\paragraph{Framework re-derivation.} Applying the feasibility set of
Eq.~(1) with $Q_{F1,\min}{=}0.85$ and $Q_{\text{cov},\min}{=}0.99$
to the seed-study means gives
$\mathcal{M}_F{=}\{\text{SmolVLM2-2.2B FT},\text{Qwen-3B FT},\text{Qwen-7B FT}\}$.
Under the illustrative case-study profile of
Section~\ref{sec:case} the cheapest survivor remains SmolVLM2-2.2B FT,
matching the $m^{*}$ derived from the main results. SmolVLM-500M FT
fails both accuracy (F1 0.832 $<$ 0.85) and coverage (0.951 $<$ 0.99)
under this resampled protocol, so its exclusion does not depend on
any single threshold.

\end{document}